\documentclass[runningheads]{llncs}
\usepackage[utf8]{inputenc}
\usepackage{amssymb}
\usepackage{amsmath}
\usepackage{float}

\usepackage[T1]{fontenc}
\usepackage{graphicx,verbatim}
\begin{document}
\title{DMD-augmented Unpaired Neural Schrödinger Bridge for Ultra-Low Field MRI Enhancement}
\titlerunning{DMD-augmented UNSB for Ultra-Low Field MRI Enhancement}

%

\author{Youngmin Kim\inst{1}\textsuperscript{*} \and
Jaeyun Shin\inst{2}\textsuperscript{*} \and
Jeongchan Kim\inst{2}\textsuperscript{*} \and
Taehoon Lee\inst{2} \and
\\ Jaemin Kim\inst{2} \and
Peter Hsu\inst{3} \and
Jelle Veraart\inst{3} \and
Jong Chul Ye\inst{2}}

\authorrunning{Y. Kim et al.}

\institute{Korea University, Seoul, Republic of Korea\\
\email{zeromin03@korea.ac.kr}
\and
KAIST, Daejeon, Republic of Korea\\
\email{\{jaeyun.shin, jchan.kim, dlxogns0128, kjm981995, jong.ye\}@kaist.ac.kr}
\and
NYU Langone Health, New York, NY, USA\\
\email{\{peter.hsu, jelle.veraart\}@nyulangone.org}
\\
\textsuperscript{*}These authors contributed equally.
}

\maketitle

\setlength{\emergencystretch}{2em}
\begin{abstract}
Ultra-Low-Field (64\,mT) brain MRI improves accessibility but suffers from reduced image quality compared to 3\,T.
As paired 64\,mT--3\,T scans are scarce, we propose an \emph{unpaired} 64\,mT$\rightarrow$3\,T translation framework that enhances realism while preserving anatomy.
Our method builds upon the Unpaired Neural Schr\"odinger Bridge (UNSB) with multi-step refinement. To strengthen target distribution alignment, we augment the adversarial objective with DMD2-style diffusion-guided distribution matching using a frozen 3T diffusion teacher. To explicitly constrain global structure beyond patch-level correspondence, we combine PatchNCE with an Anatomical Structure Preservation (ASP) regularizer that enforces soft foreground–background consistency and boundary-aware constraints.
Evaluated on two disjoint cohorts, the proposed framework achieves an improved realism–structure trade-off, enhancing distribution-level realism on unpaired benchmarks while increasing structural fidelity on the paired cohort compared to unpaired baselines.
\end{abstract}

\section{Introduction}


Ultra-Low-Field (ULF) MRI systems (e.g., 64 mT) increase accessibility by reducing infrastructure requirements compared to conventional high-field scanners such as 3T~\cite{article,article2}. However, the lower field strength leads to a low signal-to-noise ratio (SNR), which blurs anatomical structures and weakens tissue contrast. These degradations reduce the reliability of downstream clinical analysis~\cite{article,article3}.

To address this gap, there is a strong motivation to translate 64 mT scans into 3T-like images. Most prior works rely on paired supervision \cite{lucas2025multisequence}. While paired learning provides stable optimization, acquiring spatially registered 64 mT–3 T scan pairs is challenging and is further complicated by protocol variability. Unpaired translation offers a practical alternative by removing the need for correspondence \cite{seibel2024anatomicalconditioningcontrastiveunpaired}. However, unpaired models are prone to anatomical distortion or structural hallucination. In medical imaging, where decisions depend on faithful structural representation, preserving subject-specific anatomy is essential \cite{inbook}.

%


To address this limitation, we build upon the UNSB (Unpaired Neural Schrödinger Bridge) formulation \cite{kim2023unpaired} for unpaired 64mT$\rightarrow$3T translation. While UNSB enables distributional transport between domains, its alignment to the target distribution is primarily driven by a single discriminator. This adversarial signal is insufficient to capture the full complexity of 3T imaging characteristics, including subtle tissue contrast and fine-grained anatomical details.

To provide stronger target-domain guidance, we incorporate DMD2-style diffusion-guided distribution matching \cite{yin2024improved} using a teacher diffusion model pre-trained on real 3T scans. The teacher supplies informative distribution-level gradients, encouraging generated images to better align with the authentic target manifold while preserving structural details \cite{yin2024onestepdiffusiondistributionmatching}.

\section{Related Works}
%
\label{sec:bg_dmd2}


Improved Distribution Matching Distillation (DMD2) \cite{yin2024improved} introduces a diffusion-guided objective for aligning a generator distribution with a target real distribution across multiple diffusion noise levels. DMD2 leverages score differences computed from a frozen real-data diffusion teacher and a trainable fake diffusion critic, providing informative gradients.


Let $F(\cdot,\tau)$ denote the forward diffusion process at noise level $\tau$. Given a generated sample $\hat{\mathbf{y}} = G_\theta(\mathbf{x})$ produced by a generator $G_\theta$, the diffused sample at noise level $\tau$ is defined as $\mathbf u = F(\hat{\mathbf y},\tau)$. DMD2 minimizes the expected KL divergence between the distributions of noisy real and noisy generated (fake) samples at diffusion level $\tau$:
\begin{align}
\nabla_\theta \mathcal{L}_{\mathrm{DMD2}}(\theta)
&= \mathbb{E}_{\tau}\!\left[
\nabla_\theta\,\mathrm{KL}(p_{\mathrm{fake},\tau}\,\|\,p_{\mathrm{real},\tau})
\right].
\label{eq:dmd2_grad}
\end{align}

Expanding the KL divergence and applying the chain rule yields
\begin{align}
\nabla_\theta\,\mathrm{KL}\!\left(p_{\mathrm{fake},\tau}\,\|\,p_{\mathrm{real},\tau}\right)
&=
\mathbb{E}_{\mathbf u \sim p_{\mathrm{fake},\tau}}
\left[
\nabla_\theta\Big(\log p_{\mathrm{fake},\tau}(\mathbf u) - \log p_{\mathrm{real},\tau}(\mathbf u)\Big)
\right] \nonumber\\
&=
-\,\mathbb{E}_{\mathbf u \sim p_{\mathrm{fake},\tau}}
\left[
\Big(s_{\mathrm{real}}(\mathbf u,\tau)-s_{\mathrm{fake}}(\mathbf u,\tau)\Big)
\frac{\partial \mathbf u}{\partial \theta}
\right],
\label{eq:kl_grad_score}
\end{align}
where $s_{\mathrm{real}}(\mathbf u,\tau) = \nabla_{\mathbf u}\log p_{\mathrm{real},\tau}(\mathbf u)$ and
$s_{\mathrm{fake}}(\mathbf u,\tau) = \nabla_{\mathbf u}\log p_{\mathrm{fake},\tau}(\mathbf u)$.
The real score $s_{\mathrm{real}}(\cdot,\tau)$ is approximated by a frozen diffusion teacher $\mu_{\mathrm{real}}$ trained on the target real data, and
$s_{\mathrm{fake}}(\cdot,\tau)$ is approximated by a fake diffusion critic $\mu_{\mathrm{fake}}$ with parameters $\eta$ that is dynamically updated during training to track the generator's distribution  \cite{yin2024improved}.
Our goal is to synergistically incorporate DMD2 into the unpaired neural Schr\"odinger bridge 
for ULF-MRI enhancement without matched high field MRI data.

\begin{figure}[h]
\centering
\includegraphics[width=\linewidth]{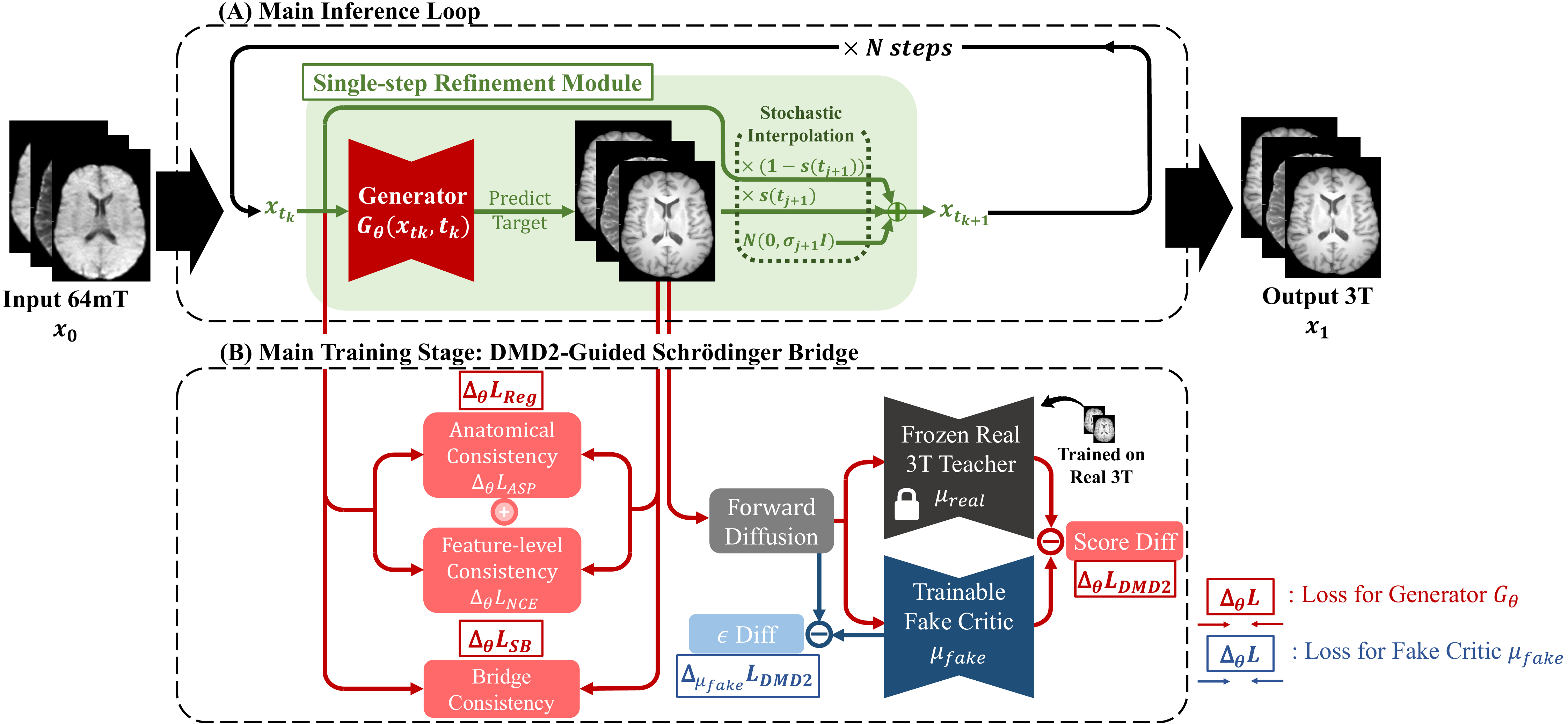}
\caption{\textbf{Overview.} The proposed framework builds upon UNSB to perform multi-step 64\,mT$\rightarrow$3\,T translation. During training, DMD2-based diffusion-guided distribution alignment using a frozen 3T teacher and a trainable fake critic augments adversarial supervision. Structural fidelity is enforced through combined PatchNCE and ASP regularization, which constrains anatomical consistency.}
\label{fig:figure1}
\end{figure}

\section{Method: DMD2-augmented UNSB}

We consider \emph{unpaired} translation from low-field 64mT brain MRI to high-field 3T MRI.
Let $\mathbf{x}\sim\mathcal{X}$ and $\mathbf{y}\sim\mathcal{Y}$ denote the domains of 64mT and 3T scans, respectively.
Each sample is represented as a 3-channel 2D axial slice to leverage a pre-trained backbone originally optimized for RGB images., and we use the same channel construction
for both source and target:
\[
\mathbf{x} = [x^{\mathrm{T1}},\,x^{\mathrm{T2}},\,x^{\mathrm{T1}}],\qquad
\mathbf{y} = [y^{\mathrm{T1}},\,y^{\mathrm{T2}},\,y^{\mathrm{T1}}].
\]



\paragraph{\textbf{UNSB Formulation.}}
\label{sec:unsb}

We  upon the Unpaired Neural Schrödinger Bridge (UNSB) \cite{kim2023unpaired}, which formulates translation as a stochastic transport between source and target distributions. The Schrödinger Bridge (SB) models this transport through a sequence of intermediate refinement states, enabling gradual distribution alignment instead of a direct one-step mapping. Such progressive refinement is particularly suitable for medical image translation, where preserving anatomical structure is critical.

Given an input low-field slice $\mathbf{x}$, we discretize the SB time interval $t\in[0,1]$ into $K$ refinement steps with a grid $0=t_0 < t_1 < \cdots < t_K=1$, and initialize the trajectory as $\mathbf{x}_{t_0} :=\mathbf{x}$. At each step $k=0,\cdots,K-1$, the generator predicts a
target-domain sample conditioned on the current intermediate state:
\begin{equation}
\hat{\mathbf{x}}_{1}^{(k)} = G_{\theta}\!\left(\mathbf{x}_{t_k},\, t_k\right).
\label{eq:unsb_pred}
\end{equation}
The next intermediate state is then obtained by stochastically interpolating between the current
state $\mathbf{x}_{t_k}$ and the predicted target sample $\hat{\mathbf{x}}_{1}^{(k)}$:
\begin{equation}
\mathbf{x}_{t_{k+1}} = (1-\alpha_k)\,\mathbf{x}_{t_k}
\;+\; \alpha_k\,\hat{\mathbf{x}}_{1}^{(k)}
\;+\; \sigma_k\,\boldsymbol{\varepsilon}_k,
\qquad \boldsymbol{\varepsilon}_k \sim \mathcal{N}(\mathbf{0},\mathbf{I}),
\label{eq:unsb_update}
\end{equation}
for $k=0,\ldots,K-1$. Here $\alpha_k := \frac{t_{k+1}-t_k}{1-t_k}$ and $\sigma_k$ follow the Gaussian bridge schedule used in UNSB, and $\boldsymbol{\varepsilon}_k$ denotes injected stochasticity (e.g., Gaussian noise) used during training.
At inference, we set $\boldsymbol{\varepsilon}_i=\mathbf{0}$ to obtain deterministic refinement and preserve anatomical consistency. 
The final translated output is given by the last target prediction, $\hat{\mathbf{y}} =\hat{\mathbf{x}}_{1}^{(K-1)}$.
For each SB refinement step $i$, UNSB optimizes the following objective:
\begin{equation}
\mathcal{L}_{\mathrm{UNSB}}(\theta,k)
:= \mathcal{L}_{\mathrm{Adv}}(\theta,k)
+ \lambda_{\mathrm{SB}}\,\mathcal{L}_{\mathrm{SB}}(\theta,k)
+ \lambda_{\mathrm{Reg}}\,\mathcal{L}_{\mathrm{Reg}}(\theta,k),
\label{eq:unsb_orig}
\end{equation}
where $\mathcal{L}_{\mathrm{Adv}}$ is the adversarial distribution-matching term, $\mathcal{L}_{\mathrm{SB}}$ is the SB loss, and $\mathcal{L}_{\mathrm{Reg}}$ is a structure-preserving regularizer using PatchNCE  \cite{kim2023unpaired}.

In this work, we retain the multi-step stochastic transport formulation of UNSB while strengthening the adversarial alignment $\mathcal{L}_{\mathrm{Adv}}$ and structural regularization $\mathcal{L}_{\mathrm{Reg}}$ for ultra-low-field to high-field MRI translation.



\paragraph{\textbf{DMD2 Augmented Adversarial Loss.}}
Instead of relying solely on a discriminator operating on the final output, we augment the adversarial objective with diffusion-guided distribution matching based on DMD2 (Sec.~\ref{sec:bg_dmd2}) at each refinement step. Specifically, given the intermediate state $\mathbf{x}_{t_k}$ at step $k$, the generator predicts the target-domain sample $\hat{\mathbf{y}}_k := \hat{\mathbf{x}}_1^{(k)} = G_\theta(\mathbf{x}_{t_k}, t_k)$, which we plug into the DMD2 gradient formulation (\ref{eq:dmd2_grad}). To steer the generated distribution toward the high-field domain $\mathcal{Y}$, the true score $s_{\mathrm{real}}(\cdot,\tau)$ is approximated by a frozen diffusion teacher $\mu_{\mathrm{real}}$ pre-trained on real 3T brain MRI data. The step-wise DMD2 objective minimizes the expected KL divergence between the diffused intermediate predictions and the diffused real 3T distribution:
\begin{equation}
\begin{aligned}
\mathcal{L}_{\mathrm{DM}}(\theta, k)
&=
\mathcal{L}_{\mathrm{Adv}}(\theta, k)
+
\mathcal{L}_{\mathrm{DMD2}}(\theta, t_k), \\[4pt]
\mathcal{L}_{\mathrm{DMD2}}(\theta, t_k)
&=
\mathbb{E}_{\mathbf{x}_{t_k},\,\tau}\!\left[
\mathrm{KL}\!\left(
p_{\mathrm{fake},\tau}
\,\big\|\,
p^{(3\mathrm{T})}_{\mathrm{real},\tau}
\right)
\right].
\end{aligned}
\tag{6}
\end{equation}
By dynamically updating the fake diffusion critic $\mu_{\mathrm{fake}}$, the generator receives score-based gradients that complement the adversarial signal and promote realistic 3T tissue contrasts throughout the refinement trajectory.

\paragraph{\textbf{Regularization for Anatomical Structure Preservation.}}
In unpaired MRI translation, enforcing only patch-level correspondence can lead to morphological artifacts, such as foreground-background leakage or boundary drift. While PatchNCE in original UNSB formulation (\ref{eq:unsb_orig}) promotes patch-level consistency, it does not explicitly constrain global foreground separation or boundary alignment. To address this limitation, we introduce an Anatomical Structure Preservation (ASP) regularizer that  constrains the generated sample to follow the foreground–background structure of the original ULF slice.

We first consider foreground preservation. Let set $\mathbf{x}_{01}=(\mathrm{clip}(\mathbf{x},-1,1)+1)/2$ as a normalized input and define the soft mask function:
\begin{equation}
m(\mathbf{x})=\sigma\left(\frac{\mathbf{x}_{01}-\tau_m}{s_m}\right),
\end{equation}
where $\tau_m$ and $s_m$ control the mask threshold and its softness, and $\sigma(\cdot)$ represents the sigmoid function. We then set $m_{\mathrm{in}}=m(\mathbf{x})$ and $m_{\mathrm{out}}=m(\hat{\mathbf{y}}_k)$, where $\hat{\mathbf{y}}_k=G_\theta(\mathbf{x}_{t_k},t_k)$. From $m_{\mathrm{in}}$, we construct a trimap with confident foreground and background regions using thresholds $\tau_{\mathrm{fg}}$ and $\tau_{\mathrm{bg}}$:
\begin{equation}
    m_{\mathrm{core}}=\mathbb{I}[m_{\mathrm{in}}>\tau_{\mathrm{fg}}], \quad m_{\mathrm{bg}}=\mathbb{I}[m_{\mathrm{in}}<\tau_{\mathrm{bg}}].
\end{equation}

For boundary preservation, we use a Normalized Surface Distance (NSD)-style boundary precision term \cite{Antonelli_2022}. Let $b_{\mathrm{out}}$ be a boundary map extracted from the output soft mask $m_{\mathrm{out}}$. We compute a distance transform $d\in\mathbb{R}^{H\times W}$ to the boundary of the input mask $m_{\mathrm{in}}$, so that $d=0$
on the boundary. The parameters $t$ and $\gamma$ control the tolerance and softness of the constraint, respectively.
We compute the ratio of the pixels of the boundary $b_{\mathrm{out}}$ that lie within distance $t$ of the input boundary via the sigmoid gate, and penalize the complement of this inlier ratio to enforce tight boundary alignment.


The ASP loss combines (i) a trimap-based mask-consistency term using weighted Binary Cross Entropy (BCE) on confident regions, and (ii) a boundary-aware constraint that penalizes edges far from the input boundary:
\begin{align}
\mathcal{L}_{\mathrm{ASP}}
&=
\underbrace{\!\left[
\mathrm{BCE}(m_{\mathrm{out}},\mathbf{1};m_{\mathrm{core}})
+\mathrm{BCE}(m_{\mathrm{out}},\mathbf{0};m_{\mathrm{bg}})
\right]}_{\text{trimap mask consistency}}
+\underbrace{\!\left[
1-\frac{\sum b_{\mathrm{out}}\,
\sigma\!\left(\frac{t-d}{\gamma}\right)}{\sum b_{\mathrm{out}}}
\right]}_{\text{boundary-aware (NSD-style)}}.
\end{align}

Finally, we combine PatchNCE with ASP to form the regularizer:
\begin{equation}
\mathcal{L}_{\mathrm{Reg}}=
\mathcal{L}_{\mathrm{PatchNCE}}
+\mathcal{L}_{\mathrm{ASP}}.
\end{equation}

\paragraph{\textbf{Overall Objective.}} 
 By combining the aforementioned components, the overall objective optimized at each SB refinement step $k$ is defined as:
\begin{align}
\mathcal{L}_{\mathrm{Total}}(\theta,k)
&= \lambda_{\mathrm{DM}}\mathcal{L}_{\mathrm{DM}}(\theta,k)
+ \lambda_{\mathrm{SB}}\,\mathcal{L}_{\mathrm{SB},k}(\theta,k)
+ \lambda_{\mathrm{Reg}}\mathcal{L}_{\mathrm{Reg}}(\theta,k),
\end{align}
where $\mathcal{L}_{\mathrm{SB}}$ denotes the standard Schr\"odinger Bridge matching loss that aligns the stochastic trajectories \cite{kim2023unpaired},
and $\lambda_{\mathrm{DMD2}}$, $\lambda_{\mathrm{SB}}$, $\lambda_{\mathrm{Reg}}$ are hyperparameters controlling the relative importance of each term.
To stabilize optimization of the DMD2 objective, we employ a two time-scale update rule (TTUR) and an auxiliary GAN classifier following DMD2 \cite{yin2024improved}. The overview of our framework is illustrated in Fig.~\ref{fig:figure1}.

\section{Experiments}

\paragraph{\textbf{Datasets.}}
\label{sec:exp_data}
We use two independent cohorts: 64mT brain MRI (Site A, $n=86$) from a public Zenodo dataset~\cite{zenodo} and 3T brain MRI (Site B, $n=181$) from the IXI Hammersmith Hospital 3\,T subset~\cite{ixi}, each providing T1w/T2w volumes. For unpaired training, we sample 64\,mT inputs from the Zenodo training split and 3\,T targets from the full IXI cohort as the target-domain pool. We evaluate on the held-out Zenodo test split ($n=5$), which provides paired 64\,mT--3\,T scans; on this set, we report full-reference paired metrics using the paired 3\,T scans and additionally compute unpaired metrics on the translated outputs, with distribution-level realism assessed against the IXI 3\,T pool.

\paragraph{\textbf{Protocol.}}
We apply subject-wise brain masking (skull stripping) to focus on the brain region. Registration is performed using ANTsPy~\cite{ants}. Within each domain, we register T2$\rightarrow$T1 per subject to ensure intra-subject consistency. For geometric normalization, we register all 3\,T volumes to a fixed 64\,mT reference subject selected from the Zenodo training split (3\,T$\rightarrow$64\,mT). 64\,mT volumes are kept in their native spaces after skull stripping. We use subject-level splits for the 64\,mT cohort (81/5 train/test) to avoid slice leakage, while the IXI cohort is used only as the target-domain pool.

\begin{table}[t]
\caption{\textbf{Quantitative evaluation.} (A) Unpaired test metrics. (B) Paired test metrics (Arrows indicate better.) \textbf{Bold}: best, \underline{Underline}: second best. }
\centering
\small
\setlength{\tabcolsep}{4pt}
\textbf{(A) Unpaired test-set evaluation}\\[1pt]
\begin{tabular*}{\linewidth}{@{\extracolsep{\fill}}lcccc}
\hline
\textbf{Method} & \textbf{Rad-FID$\downarrow$} & \textbf{FID$\downarrow$} & \textbf{KID$\downarrow$} & \textbf{PIQE$\downarrow$} \\
\hline
CycleGAN \cite{zhu2020unpairedimagetoimagetranslationusing}  & 0.3551 & 26.5839 & 0.0073 & 37.8657 \\
CUT \cite{park2020contrastivelearningunpairedimagetoimage}     & 0.3011 & 23.0661 & 0.0070 & 47.4585  \\
SDEdit \cite{meng2022sdeditguidedimagesynthesis}    & 0.5276 & 28.3546 & 0.0120 & 49.3573 \\
SynDiff \cite{özbey2023unsupervisedmedicalimagetranslation}   & 0.9415 & 46.6598 & 0.0261 & \underline{36.7951}\\
RegGAN \cite{kong2021breakingdilemmamedicalimagetoimage}    & 0.6595 & 25.2079 & 0.0048 & \textbf{34.3080}  \\
INR-Based \cite{islam2025ultra} & 1.2599 & 69.3590 & 0.0429 & 52.5423  \\
UNSB \cite{kim2023unpaired}      & 0.2688 & 19.8965 & \textbf{0.0029} & 54.9813  \\
\hline
UNSB + ASP    & 0.2596 & 20.5324 & 0.0048 & 50.7232 \\
UNSB + DMD2   & \textbf{0.2427} & \underline{19.2904} & 0.0034 & 47.1204   \\
\textbf{Ours} & \underline{0.2516} & \textbf{18.9950} & \underline{0.0031} & 48.4852\\
\hline
\end{tabular*}

\textbf{(B) Paired evaluation (independent cohort)}\\[1pt]
\begin{tabular*}{\linewidth}{@{\extracolsep{\fill}}lcccc}
\hline
\textbf{Method} & \textbf{MS-SSIM(T1)$\uparrow$} & \textbf{PSNR(T1)$\uparrow$} & \textbf{MS-SSIM(T2)$\uparrow$} & \textbf{PSNR(T2)$\uparrow$} \\
\hline
CycleGAN \cite{zhu2020unpairedimagetoimagetranslationusing}  & 0.9204 & 23.3010 & 0.8926 & 22.5735 \\
CUT \cite{park2020contrastivelearningunpairedimagetoimage}       & 0.9046 & 22.4057 & 0.8870 & 22.6232 \\
SDEdit \cite{meng2022sdeditguidedimagesynthesis}    & 0.5478 & 14.0735 & 0.5297 & 16.4094 \\
SynDiff \cite{özbey2023unsupervisedmedicalimagetranslation}   & 0.7712 & 18.3572 & 0.7166 & 18.7577 \\
RegGAN \cite{kong2021breakingdilemmamedicalimagetoimage}    & 0.7062 & 16.7101 & 0.7000 & 18.8759 \\
INR-Based \cite{islam2025ultra} & 0.8618 & 19.9516 & 0.8926 & \textbf{25.2925} \\
UNSB \cite{kim2023unpaired}      & 0.9223 & 23.4586 & 0.8952 & 22.8230 \\
\hline
UNSB + ASP   & 0.9231  &23.1991 & \underline{0.9037} & 22.7681 \\
UNSB + DMD2  &  \underline{0.9336} & \underline{23.8217} & 0.9025 & \underline{23.1992} \\
\textbf{Ours} & \textbf{0.9345} & \textbf{24.0523} & \textbf{0.9083} & 23.0436 \\
\hline
\end{tabular*}

\label{tab:table1}
\end{table}

\label{sec:results}
\begin{figure}[!h]
\centering
\includegraphics[width=\linewidth]{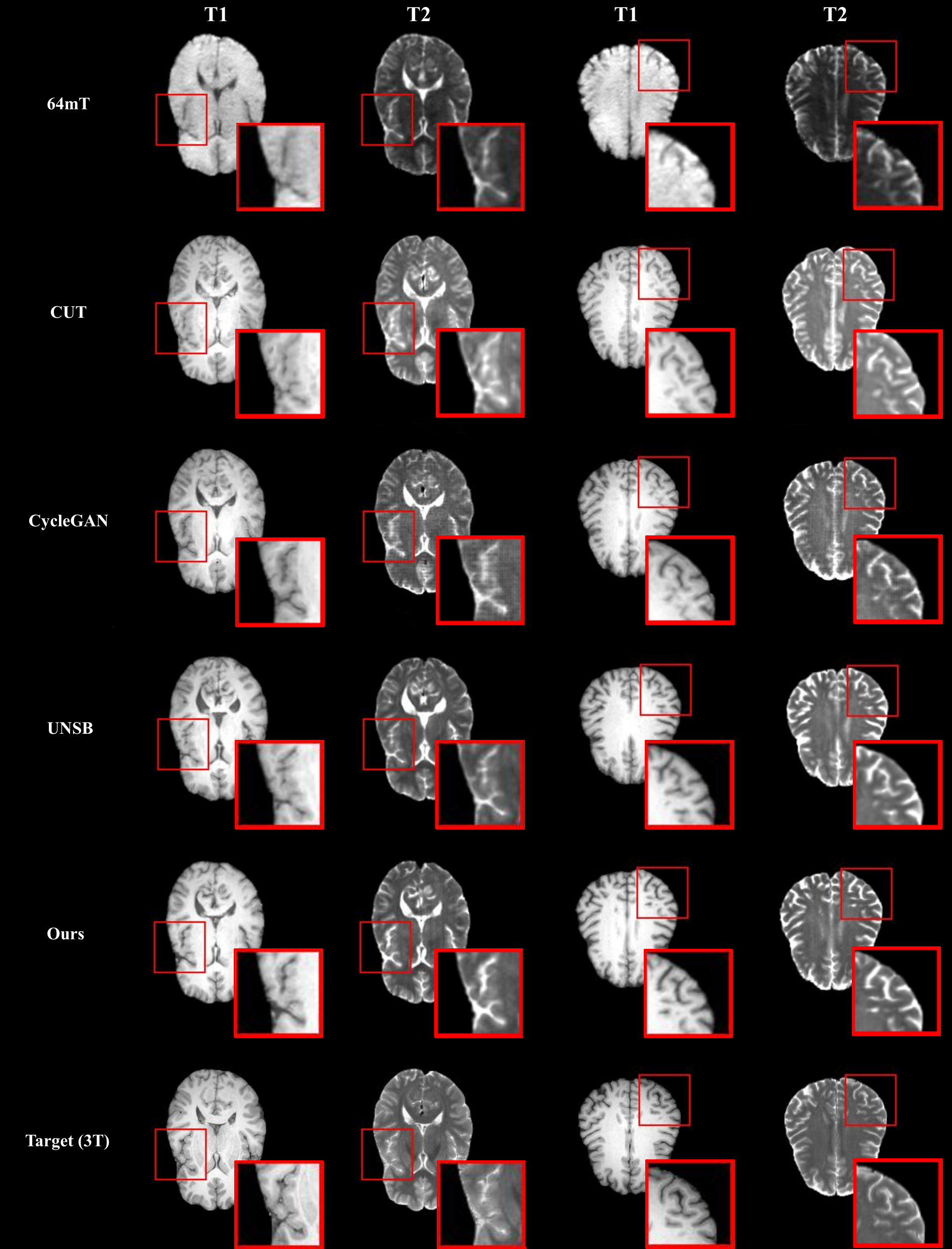}
\caption{\textbf{Qualitative comparison on the test set.} The figure shows two different slices extracted from distinct subjects within the test set. Rows: 64mT input, CUT, CycleGAN, UNSB, Ours, and Target (3T). Columns: T1/T2  and T1/T2 .}
\label{fig:qualitative}
\end{figure}

\label{sec:exp_baselines}
\paragraph{\textbf{Baselines.}}
We compare against representative unpaired translation methods, including GAN-based models:
CycleGAN \cite{zhu2020unpairedimagetoimagetranslationusing}, CUT \cite{park2020contrastivelearningunpairedimagetoimage}, RegGAN \cite{kong2021breakingdilemmamedicalimagetoimage}, and diffusion-based models: SDEdit \cite{meng2022sdeditguidedimagesynthesis}, 
SynDiff \cite{özbey2023unsupervisedmedicalimagetranslation},
INR-Based \cite{islam2025ultra}, and UNSB \cite{kim2023unpaired}.
When available, we use official implementations and apply a unified pre-processing and evaluation protocol. We also evaluate UNSB + DMD2 and other variations of our method using same hyperparameter settings.

%
%

\paragraph{\textbf{Unpaired evaluation.}}
Following the INR-based protocol~\cite{islam2025ultra}, we report no-reference perceptual quality (PIQE \cite{venkatanath2015blind}) computed within the brain mask. To quantify target-domain realism at the distribution level, we additionally report FID \cite{heusel2018ganstrainedtimescaleupdate} and KID \cite{bińkowski2021demystifyingmmdgans} in a medical feature space using a frozen 3\,T-domain encoder trained on 3\,T \emph{training} data only, as well as Rad-FID~\cite{woodland2024_fid_med} computed using a frozen radiology-domain encoder trained on 3\,T \emph{training} data, analogous to FID. All unpaired metrics are computed on the held-out unpaired test set with subject-wise aggregation.

\paragraph{\textbf{Paired evaluation.}}
Although all models are trained strictly in an unpaired manner, it is important to assess how effective the learned translation is when applied to real paired data.
To this end, we additionally evaluate full-reference fidelity on a small independent cohort of patients for whom paired 64\,mT--3\,T scans are available.
This paired cohort is disjoint from all training data and is used only for testing.
We report PSNR \cite{gonzalez2018digital} and MS-SSIM \cite{ms-ssim} between the translated outputs and the corresponding 3\,T targets.
To avoid slice-count imbalance, we first average each metric over slices per subject and then report the mean across subjects.

\section{Experimental Results}

\paragraph{\textbf{Quantitative Results.}}
\label{sec:results_quant}

We summarize the quantitative results in Table~\ref{tab:table1}. In the unpaired setting (Table~\ref{tab:table1}A), our method achieves the best FID and competitive scores on the remaining metrics, indicating improved target-domain realism without sacrificing structural consistency. On the independent paired cohort (Table~\ref{tab:table1}B), we further observe the highest PSNR and MS-SSIM on T1 and strong performance on T2. Table~\ref{tab:table1} also includes ablations on top of UNSB, showing that DMD2 strengthens target-domain alignment and that the full model provides the best overall realism--fidelity trade-off.

\paragraph{\textbf{Qualitative Results.}}
Figure~\ref{fig:qualitative} compares representative unpaired and paired test cases using full field-of-view and zoomed crops.
Our method yields sharper tissue interfaces and more realistic 3T-like textures while preserving input-consistent anatomy, whereas baselines may over-smooth or introduce artifacts or hallucinated edges near fine boundaries.

\section{Conclusion}
\label{sec:conclusion}
We proposed a structure-preserving unpaired 64\,mT$\rightarrow$3\,T MRI translation framework based on a Schr\"odinger Bridge.
By augmenting the UNSB adversarial objective with DMD2, the model achieves improved alignment to the target distribution. In addition, the ASP loss constrains foreground separation and boundary consistency, promoting faithful structural reconstruction. Extensive evaluation demonstrates improved distribution-level realism on unpaired benchmarks and enhanced full-reference fidelity on an independent paired cohort, indicating a favorable balance between perceptual realism and anatomical integrity. A current limitation is the 2D slice-based design, which does not explicitly model volumetric context, introducing inter-slice inconsistencies. Future work will extend the framework to incorporate 3D context and explicit volumetric consistency constraints.

%
%
%
\bibliographystyle{unsrt}
\bibliography{mybibliography}
\end{document}